\relax
\documentclass[letterpaper]{article} 
\usepackage{aaai22}  
\usepackage{times}  
\usepackage{helvet}  
\usepackage{courier}  
\usepackage[hyphens]{url}  
\usepackage{graphicx} 
\usepackage{natbib}  
\usepackage{caption} 
\DeclareCaptionStyle{ruled}{labelfont=normalfont,labelsep=colon,strut=off} 
\usepackage{algorithm}
\usepackage{algorithmic}

\usepackage{newfloat}
\usepackage{listings}
\floatstyle{ruled}
\newfloat{listing}{tb}{lst}{}
\floatname{listing}{Listing}

\usepackage{bibentry}

\usepackage[utf8]{inputenc} 
\usepackage{url}            
\usepackage{booktabs}       
\usepackage{amsfonts}       
\usepackage{nicefrac}       
\usepackage{microtype}      
\usepackage{xcolor}         

\usepackage{url}
\usepackage{bbm}
\usepackage{microtype}
\usepackage{graphicx}
\usepackage{subfigure}
\usepackage{booktabs} 
\usepackage{xcolor}
\usepackage{amsthm}
\usepackage{multirow}
\usepackage{diagbox}
\usepackage{arydshln}
\usepackage{mathtools}
\usepackage{soul}
\soulregister\cite7
\soulregister\gls7
\soulregister\etal7

\usepackage{booktabs}
\usepackage{hhline}
\usepackage{tabularx}
\newcolumntype{Y}{>{\centering\arraybackslash}X}
\newcolumntype{s}{>{\hsize=.8\hsize}Y}
\newcolumntype{t}{>{\hsize=.6\hsize}Y}

\newcolumntype{?}{!{\vrule width 1pt}}
\usepackage{multirow}
\usepackage{makecell}
\usepackage{dsfont}

\usepackage{commath} %

\title{On Disentangled and Locally Fair Representations}
\author{
        Yaron Gurovich, Sagie Benaim, Lior Wolf
}
\affiliations{
    School of Computer Science, Tel Aviv University
}

\begin{document}
\maketitle

%
\begin{abstract}

We study the problem of performing classification in a manner that is fair for sensitive groups, such as race and gender.
This problem is tackled through the lens of disentangled and locally fair representations. We learn a locally fair representation, such that, under the learned representation, the neighborhood of each sample is balanced in terms of the sensitive attribute. For instance, when a decision is made to hire an individual, we ensure that the $K$ most similar hired individuals are racially balanced. Crucially, we ensure that similar individuals are found based on attributes not correlated to their race. To this end, we disentangle the embedding space into two representations. The first of which is correlated with the sensitive attribute while the second is not. We apply our local fairness objective only to the second, uncorrelated, representation. Through a set of experiments, we demonstrate the necessity of both disentangled and local fairness for obtaining fair and accurate representations. We evaluate our method on real-world settings such as predicting income and re-incarceration rate and demonstrate the advantage of our method. 
\end{abstract}

\section{Introduction}
\label{sec:intro}

 Machine learning technology in domains with high social impact such as health, law, and finance, should be handled with particular care. Beyond test accuracy, one must make sure that additional social constraints are met. 
For instance, when deciding if to hire an individual, it may be the case that, for the collected data, a particular racial group exhibits higher positive outcomes resulting in an unfair model~\cite{barocas2016big,feldman2015certifying,primus2010future}. One must also ensure that the false positive and false negative rates are roughly similar across ethnic groups~\cite{zafar2017fairness}. 

In this paper, we tackle this problem through the lens of locally fair representations. Consider, for example, the task of learning a classifier for hiring an individual for a job. Our key idea is based on a simple principle: for every individual that is hired (resp. not hired), on average, the set of most similar individuals that are also hired (resp. not hired) must be racially unbiased, proportionally to their total population ratio. This gives rise to our local fairness loss. For either the hired or non-hired individuals, we consider the neighborhood of the $K$ nearest-neighbors in the feature space of our classifier. We encourage our model to make predictions such that, for example, the probability of African-American or Caucasian in this neighborhood, matches this probability in the entire population.  

One must take particular care when measuring the similarity of individuals. To give a concrete example, we want to ensure that individuals are not compared to each other based on attributes related to race, but rather on attributes such as job experience or education, which may be predictive of employability. To this end, we consider the setting of class-based disentanglement. 

Class-based disentanglement methods~\cite{lample2017fader,hadad2018two,benaim2019domain}, focus on producing two representations of the data $x$.  The first contains the information relevant to the class, and the second contains the rest of the information. This second part is a representation of the data that contains all the information that is not correlated with the sensitive attribute, such as race or gender, and it is produced without the knowledge of the downstream task and labels. In our setting, we can consider the similarity of individuals only in the representation which is not correlated to the sensitive attribute. 

Since real-world scenarios often involve  training labels that are biased with an unknown rate, learning a representation that is mostly independent on the downstream task is essential to achieve high fairness rates. Therefore,  unsupervised disentanglement has a major impact on the obtained fairness. This is evident from our experiments, which repeatedly show that the tradeoff between accuracy and fairness can be directly controlled through balancing the label-independent disentanglement loss-term and the classification loss. 

Overall, our method provides the following contributions: (1)
We introduce a novel local fairness loss, which ensures that, under the learned representation, the neighborhood of each sample is balanced in terms of the sensitive attribute. (2) We show how the  framework of disentangled representation can be used in the context of fairness. 
In this framework, two representations are learned: one that is correlated with the sensitive attribute and one that is uncorrelated. 
(3) We demonstrate that the disentangled representation is necessary for benefiting from learning locally fair representations. (4) We demonstrate the ability to control the tradeoff between accuracy and fairness on various real-world datasets. (5) Our method improves fairness, both in terms of disparate impact and equality of opportunity, while maintaining high accuracy. 
It excels especially in the regime of high accuracy levels.

\section{Related Work}

Previous work on fairness can be divided into three categories: pre-processing, in-processing and post-processing techniques. {\em Pre-processing}  techniques, such as \cite{calders2009building,edwards2015censoring,feldman2015certifying}, manipulate the training data, without modifying the training procedure. 
The advantage of the pre-processing techniques is that they require no knowledge of the downstream task. A basic version of our method, which uses disentanglement, but not local fairness, falls into this category. However, we further improve it to make use of labels.  
{\em Post-processing} techniques intervene after a model is trained. Previous work either modify the input before it enters the model~\cite{adler2018auditing,hardt2016equality,pleiss2017fairness} or modify the model's prediction~\cite{kamiran2010discrimination,menon2018cost}.

\noindent {\em In-processing} techniques modify the learning algorithm, for example, by modifying the loss function~\cite{bechavod2017learning,kamishima2012fairness}. Related to the disentanglement part of our algorithm are adversarial approaches~\cite{adel2019one,beutel2017data,wadsworth2018achieving}. ~\cite{adel2019one}, for instance, generate an embedding by applying both an adversarial objective and a classification loss. Since the labels might be correlated with the sensitive attribute,  such classification loss could reduce fairness to increase accuracy. Our disentanglement approach applies a similar adversarial objective, where the classification loss is only optionally applied to adjust the tradeoff between accuracy and fairness.

Our work crucially relies on a disentangled representation to induce a space where the similarity of individuals is based on attributes that are not correlated to the sensitive attribute. In this space,~\cite{locatello2019fairness} investigate the  usefulness of \textit{unsupervised} disentanglement approaches for making fair predictions. They show that both the optimal and empirical predictions may be unfair. Since the sensitive attribute is not known at training, an assumption is made that the sensitive attribute and the label are independent, which is not true in many real-world situations. Our approach does not make such an assumption and instead builds on the disentanglement approach of~\cite{hadad2018two}, which uses the certain attribute $a$ to create a representation  that is independent of $a$.  The authors of~\cite{creager2019flexibly} propose a disentanglement approach which does not make use of labels $y$ at training. Similarly to our method, two representations are learned from $x$, one relevant to $a$, and one irrelevant to it.
Compared to \cite{creager2019flexibly}, 
our method directly optimizes the fair representation with respect to the sensitive attribute and target classification task and thus allows the control of the tradeoff between fairness and accuracy.

Other approaches also investigate this setting proposing an additional correlation metric~\cite{tan2020learning}.  
Unlike these works, in our setting, the labels $y$ can be incorporated at training to adjust the accuracy-fairness tradeoff, and this can be done to a varying degree by adjusting the weight of corresponding classification loss. More crucially, we introduce a novel local fairness loss which improves the fairness of produced representation. 

\noindent {\em Individual fairness} is a fairness criterion in which two similar individuals should have the same classification~\cite{dwork2012fairness}. It was presented without an algorithmic way to compute similarity and only as means to evaluate a set of given labels. A follow-up work  assumes access to an oracle which can evaluate the distance between two individuals according to an unknown but fixed fair metric and show how to learn a fair classifier with a limited number of oracle queries~\cite{kim2018fairness}. In our work we consider $K$ neighbors and not the nearest one, and learn the appropriate metric by employing disentanglement. Equally important, our notion of locality, which considers the nearest neighbors out of the closest samples with a the same label is different, and so is our loss, which is based on the distribution of the protected attribute in the obtained set of neighbors. 

\section{Method}
\label{sec:method}

We begin by describing the problem formulation. We then define our disentangled model that does not use labels. We then introduce our novel local fairness loss which operates on the non-sensitive representation learned through this disentangled model. We also describe a classification loss, which can be used, when labels $y$ are available. 

\subsection{Problem Formulation}
We consider a set of data points $x \in \mathbb{R}^n$, each having an associated binary label $y \in \{-1, 1\}$ and a sensitive 
attribute $a \in \{0, 1\}$, where $0$ is the disadvantaged group and $1$ is the advantaged group. 
We are interested in training a classifier $C$ such that $\bar{y} = C(x)$,  $\bar{y}$ is accurate with respect to $y$ and also fair with respect to $a$. To this end, beyond classification accuracy, we consider multiple fairness measures.
\textbf{Demographic parity} ensures that positive and negative outcomes of $C$ occur at the same rate for both sensitive attributes: 
\begin{align}
    P(\bar{y} = c | a = 0) = P(\bar{y} = c | a = 1) ~~~~~ \forall c \in \{-1, 1\}\
\end{align}
where $\bar{y}$ is a the prediction made by $C$ for a random data point $x$, and $a$ is its sensitive attribute. 

We measure demographic parity for $c=1$, by the \textbf{Disparate Impact (DI)} measure~\cite{barocas2016big,feldman2015certifying,primus2010future}: 
\begin{align}    
    | P(\bar{y} = 1 | a = 0) - P(\bar{y} = 1 | a = 1)|
\end{align}
Demographic parity alone, however, is limited, in the case where the \textit{ground truth rates} differ. That is, when $P(y = c | a = 0) \neq P(y = c | a = 1)$, where $y$ is the ground truth label of a random data point $x$.  
\textbf{Equality of Odds} addresses disparities in the ground truth rates, and is defined as:
\begin{align}
    P(\bar{y} = 1  | a = 0, y = y_{gt}) = P(\bar{y} = 1  | a = 1, y = y_{gt}) 
    \label{eq:odds}
\end{align}
where $ y_{gt} \in \{-1, 1\}$. Equality of Odds requires both equal false positive and true positive rates between these groups. The 
\textbf{Equality of Opportunity (EO) } measure is obtained by considering the equality of odds focuses on the true positive rates only :
\begin{align}
    |P(\bar{y} = 1  | a = 0, y = 1) - P(\bar{y} = 1  | a = 1, y = 1)|
\end{align}
It ensures that the classification rates are equal across groups (having different sensitive attributes). 
We note that there is often a tradeoff between accuracy and fairness~\cite{zafar2017fairness_b}, since labels $y$ and sensitive attribute $a$ may be correlated. 
Our framework enables the control of this tradeoff. 

\begin{figure*}[t]
\centering
\begin{tabular}{c}
\includegraphics[width=0.65\linewidth]{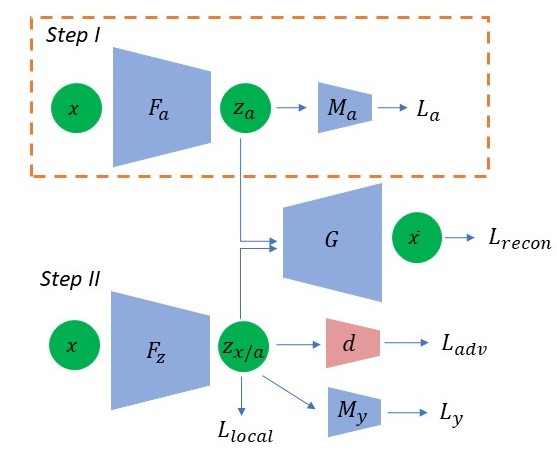}
\end{tabular}
\caption{An illustration of our method. In step I, $F_a$ is trained jointly with $M_a$ using loss $\mathcal{L}_a$, to produce representation $z_a$, which is correlated to $a$. In step II, $F_a$ and $M_a$ are frozen. We then train $F_z$ to produce a representation $z_{x/a}$ which contains the information in $x$ not correlated with $a$. To this end, $F_z$ trains, in an adversarial manner, to fool a discriminator $d$ using $\mathcal{L}_{adv}$ loss. We also apply a reconstruction loss ($\mathcal{L}_{recon}$) using a generator $G$. When labels $y$ are available, we apply the `local fairness' loss ($\mathcal{L}_{local}$) on $F_z$'s embeddings. We also train $F_z$ jointly with $M_y$ to classify label $y$. Before evaluation, we carry a final step, where $F_z$ is fixed, and $M_y$ is finetuned to minimize $L_y$. $M_y$ is then used to evaluate test samples encoded using $F_z$. }
\label{fig:illustration}
\vspace{-0.2cm}
\end{figure*}

\subsection{Disentangled Model}

Our method is illustrated in Fig.~\ref{fig:illustration}. To ensure demographic parity, it extracts a representation of $x$ that is independent of $a$ but captures all other information about $x$. 
To do so, we adopt the class-based disentanglement framework of~\cite{hadad2018two}. 
Specifically, we consider two feature extractors, $F_a$ and $F_z$, where $F_a$ encodes only the information relevant for the classification of $a$, and $F_z$ contains all other information in $x$. To this end, we train $F_a$ and $F_z$ in two steps. 

\paragraph{Step I.} We first train an MLP $M_a$ jointly with $F_a$ for classifying the correct attribute of input $x$. The following loss is optimized over $M_a$ and $F_a$:
\begin{align}
    \mathcal{L}_{a} = \frac{1}{m}\sum_{x \in X} \ell_{BCE}(M_a(F_a(x)), a) \,,
\end{align}
where $X$ is the set of training points $x \in \mathbb{R}^n$ and $a$ is the associated sensitive attribute of $x$. $m = |X|$  and $\ell_{BCE}$ stands for the binary cross-entropy loss. We choose the network architectures, such that $F_a$ maps $x$ to a low dimensional vector and $M_a$ has a low capacity, thus creating a bottleneck~\cite{tishby2015deep} that ensures that $F_a$ encodes mostly the information that is relevant to the classification of $a$.

\paragraph{Step II.} In the second step, we train $F_z$ to capture all the information in $x$ which is independent of $a$. First, we enforce that $F_z$ does not capture information about $a$. To do so, we use a domain confusion loss~\cite{tzeng2014deep} that ensures that the distribution of encoding $F_z(X_{a=0})$ equals $F_z(X_{a=1})$, where $X_{a=0}$ (resp. $X_{a=1}$) denotes the data points whose sensitive value is 0 (resp. 1). 
The loss  used is: 
\begin{align}
 \mathcal{L}_{adv} = &\frac{1}{m_0} \sum_{x_{0} \in X_{a=0} } \ell_{BCE}(d(F_z(x_0)),1) \nonumber \\ +  &\frac{1}{m_1} \sum_{x_{1} \in X_{a=1} } \ell_{BCE}(d(F_z(x_1)),1)
\end{align}
where $m_0 = |X_{a=0}|$, $m_1 = |X_{a=1}|$, 
and $d$ is a discriminator that minimizes:
\begin{align}
 \mathcal{L}_{d} = &\frac{1}{m_0} \sum_{x_{0} \in X_{a=0} } \ell_{BCE}(d(F_z(x_0)),0) \nonumber \\ + & \frac{1}{m_1} \sum_{x_{1} \in X_{a=1} } \ell_{BCE}(d(F_z(x_1)),1)
\end{align}
The discriminator $d$ attempts to separate between the distributions $F_z(X_{a=0})$ and $F_z(X_{a=1})$, by classifying samples of the former as $0$ and the samples of the latter as $1$, whereas $F_z$ tries to fool the discriminator, by trying to force the discriminator to label the embeddings $F_z(X_{a=0})$ and $F_z(X_{a=1})$ as $1$. 
Second, we use a loss term to ensure that $F_z$ captures the information in $x$ that is not captured by $F_a$. To do so, we freeze $F_a$ and require that  $F_a$'s encoding of $x$ (the factors dependent on $a$) and $F_z$'s encoding of $x$  (the factors independent of $a$) together would suffice to reconstruct $x$. That is:
\begin{align}
 \mathcal{L}_{rec} = \frac{1}{m} \sum_{i=1}^{m} \| G(F_a(x_i),F_z(x_i)) - x_i \|_1\,,
\end{align}
where $G$ is a decoder that accepts a concatenation of $F_a$'s and $F_z$'s embeddings. 

The disentanglement model alone does not make use of the labels $y$. This is useful in the case where the labels are unknown at first, and as shown in our experiments, fairness is improved through the use of these label-independent embeddings. This is a direct result of the labels being biased with respect to the sensitive group. 

\subsection{Local Fairness}

Given our disentangled representation, we can now consider our notion of `local fairness'. 
As a motivating example, consider the following setting: $x$ denotes a person's employment profile (such as education and experience). $y$ denotes a recommendation to either hire ($y=1$) or not hire ($y=-1$) this person for a given job and $a$ denotes the person's race -- African-American for $a=0$ and Caucasian for $a=1$. 
Assume that a person $x$ is not hired ($y=-1$). We would then like that the group of most similar persons to $x$ (in the embedding space), who are also not hired, to be attribute-balanced. That is, the proportion of Caucasian to African-American persons in this group should be close to their true proportion in the population. If this is not the case, then locally in the data, there is a region in which the advantageous label indicates the sensitive attribute. A similar reasoning also holds for the case of $y=1$. To obtain local neighborhoods that are apriori (before considering $y$) balanced with respect to the sensitive information, we want to ensure that similarity to other persons is done only based on information that is independent of the sensitive attribute.

To make the embedding $F_z$'s locally fair, we consider the embeddings  $F_z(X_{y=-1})$, where $X_{y=-1}$ is the set of training points for which $y=-1$ (we denote by an application of a function to the set obtained by applying this function to all set elements). Using the framework introduced earlier, $F_z(X_{y=-1})$ are learned to be non-correlated with the sensitive information and to  capture the rest of the information.  
For each embedding point $z_i \in F_z(X_{y=-1})$, we take the $K$ nearest neighbors 
in this embedding space, denoted by $z^1_i, \dots, z^K_i$.  

Assume the ratio of the protected group ($a=1$) to the non-protected group ($a=0$) in the population is $1:r$. We want that, on average, $1/(r+1)$ of the $z^1_i, \dots, z^K_i$ to have $a=0$ and $a=1$ otherwise. Let $a'^j_i$ be $-1$ if its associated sensitive attribute ($a$) is $0$, and otherwise set $a'^j_i$ to be $r$. We enforce this requirement by setting that $\| a'^1_i \cdot z^1_i + \dots + a'^K_i \cdot z^K_i \|_2 = 0$.

The same loss is applied also in the case where $y=1$, in this case considering nearest neighbors from the group $F_z(X_{y=1})$. Let $m' = |X_{y=y_{gt}}|$, the local loss is defined as: 
\begin{small}
\begin{align}
\mathcal{L}^{y_{gt}}_{local} &=
 \frac{1}{m'} \sum_{z_i \in F_z(X_{y=y_{gt}})}  \| a'^1_i \cdot z^1_i + \dots + a'^K_i \cdot z^K_i \|_2 %
\label{eq:local_fairness}
\end{align}
\end{small}
We then have $\mathcal{L}_{local} =  \mathcal{L}^{0}_{local} + \mathcal{L}^{1}_{local}$.

As an intuition, consider the case where the dimensionality of the $z_i$'s is $1$ and $r=1$, that is, the ratio of the sizes of the protected to the non-protected groups in the population is $1$. 
Here, the constraint is satisfied ($\mathcal{L}_{local}=0$) if the group of $z_i$'s with $a=0$ have the same average value as the group of $z_i$'s with $a=1$. If the number of individuals in both groups is the same, then the individuals should be, on average, similar. 

On the other hand, if the non-protected group consists of many individuals, i.e., there are $M>>1$ individuals $z_i'$ for which $a=0$,  and there is a single individual $z_i''$ for which $a=1$, then in order for $\mathcal{L}_{local}$ to become zero, the embedding value $z_i''$ has to be very far from the average $z_i'$  (for instance $z_i''=M$ and $z_i'=1$).

In our method, the local fairness is added to the loss terms that are minimized as part of step II training the disentangled model. This makes the disentanglement supervised since label information is used. Furthermore, we add to this step the training of an MLP classifier $M_y$, which is trained in conjugation with $F_z,G$ (and $d$). The classification loss used to train $M_y$ is given by
\begin{align}
    \mathcal{L}_{y} = \frac{1}{m}\sum_{x \in X} \ell_{BCE}(M_y(F_z(x)), y)\,, \label{eq:cls_y}
\end{align}
and the full objective of step II that is minimized over $M_y, F_z, G$ is given by:
\begin{align}
 \mathcal{L}_{full} = \mathcal{L}_{rec} + \lambda_1 \cdot \mathcal{L}_{adv} + \lambda_2 \cdot \mathcal{L}_{local} + \lambda_3 \cdot \mathcal{L}_{y} \label{eq:full_formulation}
\end{align}
Where $\lambda_1, \lambda_2, \lambda_3$ are hyperparameters. 
$d$ is trained simultaneously to minimize $\mathcal{L}_{d}$.   As shown in the experiments, adjusting $\lambda_3$ results in controlling the tradeoff between accuracy and fairness. Using $\lambda_3=0$ leads to the maximal fairness, but the classifier, in this case, has the lowest accuracy since the representation resulting from $F_z$ is not optimized for classification accuracy. 

\paragraph{The final classifier}
To allow the same framework for different values of $\lambda_3$, and to allow the classifier to converge after the adversarial training has been stopped, the final classifier is trained after $F_z$ has been learned.
$M_y$ is finetuned to minimize $\mathcal{L}_{y}$ of Eq.~\ref{eq:cls_y} where $F_z$ is fixed and is no longer trained. To evaluate our method, test samples are first encoded using $F_z$, and then classified using $M_y$. 

\subsection{Training and Implementation Details}
\label{sec:training_details}

$F_a$ and $F_z$ are MLPs with 2 hidden layers with $(10, 20)$ hidden units, where each layer is followed by a Leaky-Relu (slope $0.2$) and Dropout (probability $0.5$). The input dimension is data-dependent and the output embedding dimension is $20$. Decoder $G$ has the reverse architecture.  
MLP's $M_a$, $M_y$ and $d$ have similar architecture with an output to which a sigmoid activation is applied. Input $x$ is pre-processed and normalized
as in CTGAN~\cite{xu2019modeling}, where discrete values are represented as one-hot vectors and continuous columns are normalized by a distribution representation and scale to $[0,1]$. We use Adam optimizer with parameters $\beta_1 = 0.5$ and $\beta_2 = 0.999$. We start with a learning rate of $0.001$ and multiply it by 0.1 every 30 epochs, with a batch size of $64$ and run for $100$ epochs. 
For every iteration of $F_z, M_y$ and $G$, we train $d$ for $20$ iterations.
Data splits were made as in the dataset description (if test and train were available), or by a random split into train/test sets. All random sampling experiments run in 5 folds.

The Local Fairness hyperparameters are set to $K = 4$.
Hyperparameters for the loss aggregation are set to $\lambda_1=1$, $\lambda_2=1$ and $\lambda_3=0.1$. These parameters are set for all datasets and were found based on the validation set of the synthetic dataset as described in the experiments section. 
The various experiments were either run on the CPU (small datasets) or a single NVIDIA GTX1080 GPU (bigger datasets).

\section{Experiments}
\label{sec:results}

We begin by evaluating our method on the real world datasets of Adult~\cite{adult} and COMPAS~\cite{compas} that have been previously used for studying fairness. We then demonstrate the tradeoff between accuracy and fairness for two additional datasets of Bank~\cite{bank} and Communities and Crime~\cite{communities}. 
We then study the necessity of both the disentangled representation and local fairness on a synthetic dataset.  We show that the maximal advantage occurs when learning the representation using both disentanglement and local fairness, as well as the standard classification loss. Lastly we evaluate the effect of our losses on accuracy and fairness when varying $K$, the local neighborhood size, as well as the bias level of the training set.

\subsection{Real World Datasets}

\begin{figure*}[t]
\centering
\begin{tabular}{@{}c@{~}c@{}}

\includegraphics[width=0.45\linewidth]{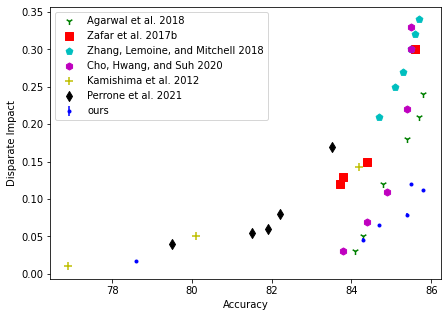} &
\includegraphics[width=0.442\linewidth]{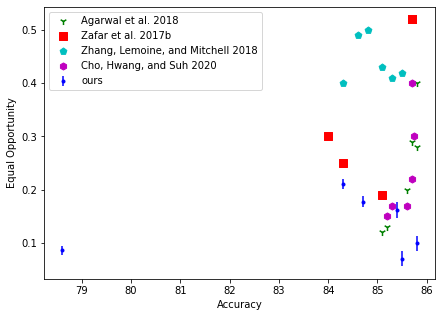} \\
(a) & (b)\\
\includegraphics[width=0.45\linewidth]{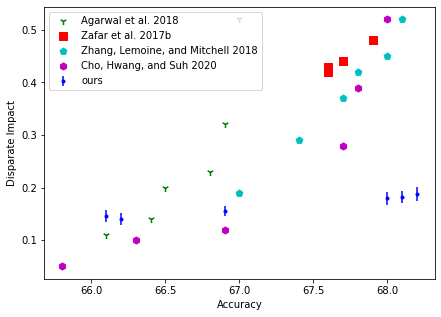} &
\includegraphics[width=0.45\linewidth]{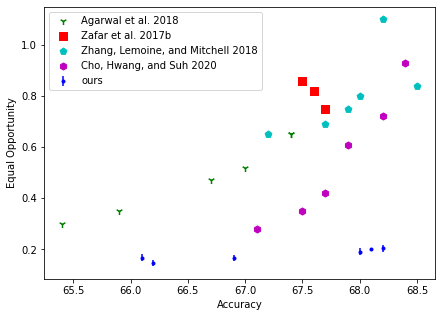} \\
(c) & (d) \\
\end{tabular}
\vspace{-2mm}
\caption{
Accuracy vs. fairness tradeoff, obtained by varying $\lambda_3$. The blue points correspond to different values of $\lambda_3$ of our method: $0, 0.1, 0.2, 0.5, 0.75, 1$, and appear from left to right in increasing accuracy (error bars indicate standard error). 
Other colors correspond to baselines with their corresponding tradeoffs. For accuracy, {higher} is better and for fairness {lower} is better. (a) Adult Disparate Impact, (b) Adult Euqality of Oppotunity. (c,d) same for COMPAS. 
}
\label{fig:adult_y}
\end{figure*}

\begin{figure*}[t]
\centering
\begin{tabular}{c}
\includegraphics[width=0.55\linewidth]{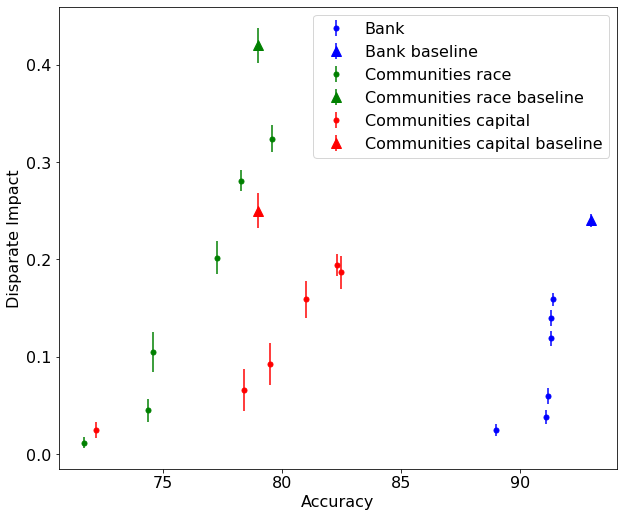} 

\end{tabular}
\caption{
Accuracy vs fairness (Disparate Impact) tradeoff for the datasets of  Bank~\cite{bank} (blue) and  Communities and Crime~\cite{communities} dataset (green+red). In the green plot, race is used as the sensitive attribute and in the red plot, the capital attribute is used. $\lambda_3 = 0, 0.1, 0.2, 0.5, 0.75, 1$, and appear from left to right in increasing accuracy in each plot. A triangle indicates a vanilla classifier trained with $\mathcal{L}_y$ only. Error bars indicate the standard error.  
}
\label{fig:bank_y}
\end{figure*}

The \textbf{Adult} dataset~\cite{asuncion2007uci} contains $32,561$ individuals with $14$ attributes, and we are asked to classify whether a given person makes more or less than \$50,000 a year; gender is the sensitive attribute. 
The \textbf{COMPAS} dataset~\cite{compas} contains 5,278 instances with 12 features, and the associated task is to predict if a criminal defendant will be re-incarcerated within two years. Here, race is the sensitive attribute. 
We compared against a set of recent baseline methods: 1. \cite{agarwal2018reductions}, 2.  \cite{zafar2017fairness_b}, 3. \cite{zhang2018mitigating}, 4. \cite{cho2020fair}, 5. \cite{kamishima2012fairness}, 6. \cite{perrone2021fair}. 
We consider the results reported in these papers (consistent with our definition of Disparate Impact and Equal Opportunity). We note that some baselines only report values on the Adult dataset for Disparate Impact.

Fig.~\ref{fig:adult_y} reports accuracy and fairness values for different weights of $\mathcal{L}_y$ in comparison to baselines for Adult and COMPAS. 
We consider both Disparate Impact and Equal Opportunity with respect to accuracy. 
For the Adult dataset, our method achieves the best tradeoff between accuracy and Disparate Impact.
Further, for maximal accuracy, it achieves the best fairness measures, both in terms of Disparate Impact and Equality of Opportunity. 
Note how in this dataset, accuracy benefits significantly from the addition of the target labels to the training and most of the gain achieved with $\lambda_3=0.1$. 

For the COMPAS dataset our method also achieves the best tradeoff between accuracy and fairness. For the highest accuracy value, the best Disparate Impact is achieved. With regards to Equal Opportunity, while our method does not achieve the overall best accuracy among baselines, it provides better fairness results for a range of accuracy values. 

\subsection{Accuracy vs Fairness Tradeoff}

We now demonstrate that the tradeoff between accuracy and fairness can be controlled using our method. To this end we consider two additional datasets. 
The \textbf{Bank} dataset~\cite{bank} contains a total of 41,188 subjects, each with 20 attributes (such as marital status) and a binary label, which indicates whether the client has subscribed (positive class) or not (negative class) to a term
deposit. We consider age as a sensitive attribute, and binarize the age attribute by assigning it to 1 if a client’s age is between 25 and 60, and 0 otherwise.
The \textbf{Communities and Crime}~\cite{communities} dataset contains 1,994 instances and 128 attributes gathered from different communities in the United States. 
The attributes are related to several factors that can highly influence some common crimes such as robberies, murders, or rapes. 
The classification task is to predict violent communities.  We take the $70th$ percentile of the total number of violent crimes per 100K population communities, as a threshold for the violence label. We consider two sensitive attributes: 'racePctBlack' (\% neighborhood population which is Black) and 'blackPerCap' (avg per capita income of Black residents).

Fig.~\ref{fig:bank_y} reports the tradeoff between accuracy and Disparate Impact on these datasets. This tradeoff is controlled through varying $\lambda_3$. As can be seen, increasing the value $\lambda_3$ results in a better accuracy and worse fairness (Disparate Impact). 
For the Communities and Crime dataset (Fig.~\ref{fig:bank_y} (green+red)), we see a large variation in Disparate Impact and accuracy, while for the Bank dataset (Fig.\ref{fig:bank_y} (blue)), we see a small accuracy variation, with significant variation in Disparate Impact. Interestingly, on the Communities and Crime dataset, high values of $\lambda_3$ provide accuracy that can reach, or even surpass, the accuracy of the vanilla classifier that employs only the $\mathcal{L}_y$ loss, while still having somewhat lower DI.

\subsection{Disentangled and Locally Fair Representations}

To demonstrate the impact of our disentangled framework and local fairness we use a synthetic dataset, which follows \cite{zhang2018mitigating}. 
Samples $(x_i, y_i, z_i)^n_{i=1}$ are generated as follows: For each sample $i$, we first generated two random seeds, $r_i, yo_i \in \{0, 1\}$. 
We then sample $v_i\sim \mathcal{N}(r_i,1)$ and $u_{r_1}, \dots, u_{r_d}\sim \mathcal{N}(v_i,1)$  (dependently on $v_i$).
We also sample $w_i\sim \mathcal{N}(v_i, 1)$ dependently on $v_i$.  Next we sample 
$vo_i\sim \mathcal{N}(yo_i,1)$ and $u_{yo_1}, \dots, u_{yo_d}\sim \mathcal{N}(vo_i,1)$ dependently on $vo_i$. 
Each sample $i$ is then defined as follows: $x^{(i)} = (r_i,u_{r_1}, \dots, u_{r_d}, u_{yo_1}, \dots, u_{yo_d})$, $y_r^{(i)} = 
\mathbbm{1}[w_i > 0]$, $y_o^{(i)} = yo_i$ and $z^{(i)} = r$ (where $\mathbbm{1}$ denotes an indicator function).
To construct the final dataset, we define $p_z$ to be the portion of biased samples in the dataset and choose the final labels $Y$ such that $p_y$ portion of the $n$ samples comes from the biased distribution $y_r^{(i)}$ and the rest from $y_o^{(i)}$.
For this experiment, we set $p_z = 0.5$ for train data and $p_z = 0$ for test data. This is done to demonstrate a case where the training phase is exposed to biased data ($50\%$ of samples) and our goal is to learn a fair predictor. In our setting, $d=25$. We use $1000$ (resp. 500) train (resp. test) samples. 

\subsubsection{Variants of Our Method}

In Tab.~\ref{tab:ablation}, we consider different variants of our method as well as of a baseline method with no disentanglement. We evaluate the ability of the resulting representation $F_z$ to predict the unbiased labels $y$ and the sensitive attribute $a$. 
We would like to get a high accuracy for the labels prediction and low accuracy (close to the random $0.5$) for the sensitive attribute.

Settings (1-4) do not contain the two-step disentanglement architecture and are similar to our 2nd step of training but without the reconstruction loss. Setting (1) uses a standard baseline classifier with only $\mathcal{L}_y$ as a loss. Setting (2) is a classifier trained with $\mathcal{L}_y$ and an adversarial loss $\mathcal{L}_{adv}$.  This is similar to \cite{adel2019one}. 
In (3) and (4) we consider these variants with our local fairness loss ($\mathcal{L}_{local}$). 
As can be seen, classifier (1) is very biased, archiving $67\%$ accuracy on the sensitive variable, and only $57\%$ accuracy on the true labels $y$ of the test set (recall that the train and test distributions differ). Adding an adversarial loss ($\mathcal{L}_{adv}$) in (2) improves the bias of the representation by reducing the accuracy to sensitive variable to $55\%$, while also improving the prediction accuracy to the true test labels $y$. Adding local fairness ($\mathcal{L}_{local}$) in (3) and (4) reduces both results towards random chance. This indicates that the classifier is unable to learn and gives a random output, showing that in the standard setting, without disentanglement, local fairness does not improve fairness. 

Settings (5-8) consider the cases where the reconstruction loss, $\mathcal{L}_{recon}$, is used, but where $\mathcal{L}_{adv}$ is not used. In these settings, the accuracy of the sensitive attribute is very high ($67-70\%$). Accuracy of the true test labels $y$ set are still not optimal ($56-59\%$). 

Settings (9-16) are based on the disentanglement framework we present. In these settings ,$\mathcal{L}_{recon}$ and $\mathcal{L}_{adv}$ are used throughout.  Using only the reconstruction ($\mathcal{L}_{recon}$) loss and adversarial loss ($\mathcal{L}_{adv}$) in (9), which is the baseline method of \cite{hadad2018two}, the sensitive accuracy is $58\%$ and the label prediction is $60\%$, which shows that disentanglement improves upon the baseline classifier. Compared to the adversarial setting of (2), it classifies the sensitive attributes more accurately (worse) but also classifies the true test labels $y$ more accurately (better). 
Adding a small portion of $\mathcal{L}_{y}$ in (10), by using $\lambda_3 = 0.1$, leaves the results the same. 
However, the significant gain is achieved by adding to (9) the local fairness loss ($\mathcal{L}_{local}$) as shown in (11). Here, the accuracy of the sensitive variable is reduced by $5\%$ to $53\%$, while leaving the label prediction rate the same.
Finally, considering, the full loss $\mathcal{L}_{full}$ by using $\lambda_1 = 1$, $\lambda_2 = 1$, $\lambda_3 = 0.1$ in (12),  achieves the best results with sensitive accuracy of $53\%$ and true labels $y$ accuracy of $61\%$. Overall, we observe that local fairness improves accuracy, but only in the case where a disentangled representation is used. 

Next, we explore the impact of $\lambda_3$ on results (classifiers (12) - (16)). Note that as $p_z$ is $0.5$ for train and $0$ for test, the proportion of biased samples is different at train and test, and so increasing $\lambda_3$ does not imply an increase in the test label $y$ accuracy. 
As $\lambda_3$ increases,  the prediction rate of the sensitive attribute also increases, as a result of a bigger exposure 
of the learning scheme to the inherent bias within the training labels. We also see that the impact on the accuracy of the true labels $y$ is reaching a plateau at $\lambda_3 = 0.1$, which indicates the limit of the relevant information in the training labels for the true (unbiased) test labels. In real-world datasets where the test set is biased, this increase in accuracy does not plateau, as our real dataset experiments show.

\begin{table}
\begin{center}
\begin{tabular}{llllllcc}
\toprule
  & $\mathcal{L}_{rec}$ & $\mathcal{L}_y$  & $\mathcal{L}_{adv}$  & $\mathcal{L}_{local}$ & Acc & Acc \\
    & & & & & $a$ $\downarrow$ & $y$ $\uparrow$ \\
   \midrule
  (1)& No & Yes ($1$) & No & No & $67\%$ & $57\%$ \\
  (2)& No & Yes ($1$) & Yes & No & $55\%$ & $58\%$ \\
  (3)& No & Yes ($1$) & Yes & Yes & $50\%$ & $50\%$ \\
  (4)& No & Yes ($1$) & No & Yes & $50\%$ & $50\%$ \\
  \midrule
  (5)& Yes& No & No & No & $67\%$ & $58\%$ \\
  (6)& Yes& Yes ($0.1$) & No & No & $66\%$ & $59\%$ \\
  (7)& Yes & No & No & Yes & $66\%$ & $56\%$ \\
  (8)& Yes & Yes ($0.1$) & No & Yes & $70\%$ & $58\%$ \\
  (9)& Yes& No & Yes & No & $58\%$ & $60\%$ \\
  (10)& Yes& Yes (0.1) & Yes & No & $58\%$ & $60\%$ \\
  (11)& Yes& No & Yes & Yes & $53\%$ & $60\%$ \\
  (12)& Yes & Yes ($0.1$) & Yes & Yes & \textbf{$53\%$} & \textbf{$61\%$} \\
  (13)& Yes& Yes ($0.2$) & Yes & Yes & $54\%$ & $61\%$ \\
  (14)& Yes& Yes ($0.5$) & Yes & Yes & $55\%$ & $61\%$ \\
  (15)& Yes& Yes ($0.75$) & Yes & Yes & $56\%$ & $60\%$ \\
  (16)& Yes& Yes ($1$) & Yes & Yes & $56\%$ & $60\%$ \\
\bottomrule
\end{tabular}
\caption{Exploring the contribution of various loss terms. 
Row (12) corresponds to our complete method. 
Row (2) corresponds to the method of \cite{adel2019one} (no disentanglement). All four variants of (1-4) include classification loss $\mathcal{L}_{y}$. Row (1) shows a regular classifier, (2) adds the adversarial loss $\mathcal{L}_{adv}$,  rows (3) and (4) add our local fairness $\mathcal{L}_{local}$.
Rows (5-8) consider the setting where $\mathcal{L}_{rec}$ is used, but where $\mathcal{L}_{adv}$ is not. 
In the disentangled setting, rows (9-16)
all rows include reconstruction loss $\mathcal{L}_{rec}$ and the adversarial loss $\mathcal{L}_{adv}$. 
(9) applies the method of \cite{hadad2018two}, focusing only on the disentanglement component. (10) adds the classification loss $\mathcal{L}_{y}$ with $\lambda_3 = 0.1$ and row (11) shows results with local fairness $\mathcal{L}_{local}$ without $\mathcal{L}_{y}$.
Rows (12- 16) show the impact of $\lambda_3$ together with $\mathcal{L}_{rec}$, $\mathcal{L}_{adv}$ and $\mathcal{L}_{local}$, by assigning $\lambda_3$ to the values $\{0.1,0.2,0.5,0.75,1\}$. The value of $\lambda_3$ is indicted in brackets. 
}
\vspace{-0.7cm}
\label{tab:ablation} 
\end{center}
\end{table}

\subsubsection{Varying the local neighborhood size}

Next we consider the effect of varying the local neighborhood size, $K$
as defined in Eq.~\ref{eq:local_fairness}. We use the synthetic dataset defined earlier, and consider three different setting: (1) $\lambda_3 = 0$, only disentanglement and local fairness is used, (2) $\lambda_3 = 0.1$, our default setting, (3) $\lambda_3 = 1$. As shown in Tab.~\ref{tab:ablation_k}, a value of $K=4$ results in the lowest (best) accuracy of the sensitive attribute $a$. A large value of $K$ (16) results in a significantly worse accuracy value of $a$, as the effect of our loss diminishes. The test label $y$ accuracy values are very similar, varying by at most $1\%$ for each setting. 

\begin{table}
\begin{center}
\begin{tabular}{cccc}
\toprule
$\lambda_3$ ($\mathcal{L}_{y}$) & $K$  & Accuracy $a$ $\downarrow$ & Accuracy $y$ $\uparrow$  \\
  \midrule
  $0$ & $2$ & $53\%$ & $59\%$\\
  $0$ & $4$ & $52\%$& $58\%$\\
  $0$ & $8$ & $53\%$& $58\%$\\
  $0$ & $16$ &$54\%$ & $59\%$\\
   \midrule
  $0.1$ & $2$ & $52\%$ & $60\%$ \\
  $0.1$ & $4$ & $52\%$ & $60\%$ \\
  $0.1$ & $8$ & $52\%$ & $60\%$ \\
  $0.1$ & $16$ & $57\%$ & $60\%$\\ 
  \midrule
  $1$ & $2$ & $56\%$ & $59\%$\\
  $1$ & $4$ & $55\%$& $59\%$\\
  $1$ & $8$ & $55\%$& $60\%$\\
  $1$ & $16$ & $58\%$& $60\%$\\
\bottomrule
\end{tabular}
\caption{Effect of varying $K$, the local neighborhood of samples considered as part of the local fairness loss. We consider three different settings with $\lambda_3=0$ (only disentanglement and local fairness), $\lambda_3=0.1$ (our default setting) and $\lambda_3=1$. 
}
\label{tab:ablation_k} 
\end{center}
\vspace{-0.6cm}
\end{table}

\begin{figure*}
\centering
\begin{tabular}{cc}
\includegraphics[width=0.45\linewidth]{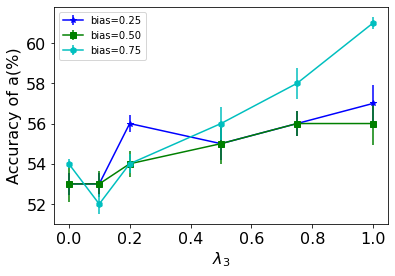} & 
\includegraphics[width=0.45\linewidth]{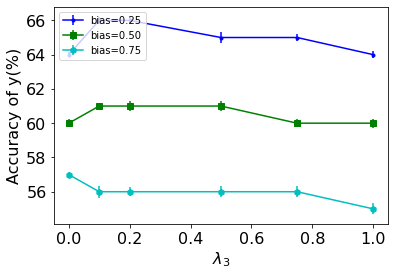} \\
(a)  & (b) \\
\end{tabular}
\caption{Effect of varying the training set bias $p_z$ (synthetic dataset) on the accuracy of the sensitive attribute (a), where lower is better, and accuracy of the test labels $y$ (b), where higher is better. Note the labels prediction $y$ is measured on test data for which $p_z = 0$, that is no biased exists. Error bars indicate the standard error.}
\label{fig:bias}
\end{figure*}

\subsubsection{Varying the bias}

We consider the effect of varying the bias, $p_z$, of the training set.  We consider the synthetic dataset as defined earlier, but where $p_z=0.25$, $p_z=0.5$ (default) and $p_z=0.75$. As can be seen in Fig.~\ref{fig:bias}(a), increasing the value of $\lambda_3$ beyond $0.1$, increases the accuracy value of the sensitive attribute a. The increase is most significant for a large bias value of $0.75$, where the accuracy reaches a value of $61\%$. With regards the the test label $y$ accuracy in Fig.~\ref{fig:bias}(b), increasing the value of $\lambda_3$ beyond $0.1$, results in either the same or lower accuracy. A value of $\lambda_3 = 0.1$ results in maximal accuracy for $p_z=0.5$ (default) and $p_z=0.25$. 

\subsubsection{Limitations}
\label{sec:limitations}

Our method has some limitations, first with respect to the ability to support multiple sensitive attributes at once. One may expand the loss terms to serve as individual heads. However, this also opens the sub-group fairness question, not addressed in this study
We believe our fairness loss can support sub-groups, but this is left for followup research.

\section{Conclusion}

In this work, we considered the role of disentangled and locally fair representations in making fair predictions. We proposed to learn a locally fair representation that ensures the neighborhood of each sample is balanced in terms of the sensitive attribute. To find this neighbourhood, we embedded samples into a feature space where samples are not correlated to the sensitive attribute. This was done by learning a disentangled representation which separates each sample into features that are correlated to the sensitive attribute and those that are not. We considered several real-world datasets, where we analyzed the trade-off between accuracy and fairness. 
For both the COMPAS and adult dataset, our method achieves the overall best tradeoff between accuracy and fairness. In particular, where high level of accuracy is required, best fairness is achieved. 
On a synthetic dataset, we demonstrated that both disentanglement and local fairness are required for fairness improvement and that a classifier further helps to control the tradeoff between accuracy and fairness. 
As future work, one could consider our notion of local fairness in the context of causal based fairness and individual fairness. 

\section{Acknowledgments}

This project has received funding from the European Research Council (ERC) under the European Union’s Horizon
2020 research and innovation programme (grant ERC CoG
725974).

\bibliography{main.bib}

\end{document}